\pdfoutput=1

\documentclass[11pt]{article}
\usepackage{url}
\usepackage[]{acl}
\usepackage{graphicx}
\usepackage{times}
\usepackage{latexsym}

\usepackage[T1]{fontenc}

\usepackage[utf8]{inputenc}
\usepackage{amsmath}
\usepackage{amssymb}
\usepackage{enumerate}
\usepackage{color}
\usepackage{CJKutf8}
\usepackage{subfigure}
\usepackage{microtype}

\newcommand\Mark[1]{\textsuperscript#1}
\title{KE-QI: A Knowledge Enhanced Article Quality Identification Dataset}

\author{Chunhui Ai\Mark{1}, Derui Wang \Mark{2}, Xu Yan \Mark{1}, Yang Xu \Mark{2} ,
Wenrui Xie \Mark{2} , Ziqiang Cao\Mark{1} \\
\Mark{1} Institute of Artificial Intelligence, Soochow University,Suzhou, China\\
\Mark{2} Baidu Inc., Beijing, China
\\
  \texttt{\{20215227120, 20215227006\}@stu.suda.edu.cn,} \\
  \texttt{\{wangderui,xuyang24,xiewenrui01\}@baidu.com,} \\
  \texttt{zqcao@suda.edu.cn}
}

\begin{document}
\maketitle
\begin{abstract}

With so many articles of varying qualities being produced every moment, it is a very urgent task to screen outstanding articles and commit them to social media.
To our best knowledge, there is a lack of datasets and mature research works in identifying high-quality articles.
Consequently, we conduct some surveys and finalize 7 objective indicators to annotate the quality of 10k articles.
During annotation, we find that many characteristics of high-quality articles (e.g., background) rely more on extensive external knowledge than inner semantic information of articles.
In response, we link extracted article entities to Baidu Encyclopedia, then propose \textbf{K}nowledge  
\textbf{E}nhanced article \textbf{Q}uality \textbf{I}dentification (KE-QI) dataset.
To make better use of external knowledge, we propose a compound model which fuses the text and external knowledge information via a gate unit to classify the quality of an article.
Our experimental results on KE-QI show that with initialization of our pre-trained Node2Vec model, our model achieves about 78\% $F_1$, outperforming other baselines.

\end{abstract}
\section{Introduction}

With the development of social networks, there are a vast amount of articles of different quality published in online media every moment, and it is essential for online platforms to automatically filter high-quality articles and distribute them to users.
We prove on our App \footnote{Baidu Feed.}, which has hundreds of millions of users, that high-quality articles effectively increase readers' interest in reading.
By presenting only outstanding articles to readers, both average reading time spent on per page and daily active users have increased obviously.

\begin{figure}[tb] 
    \centering 
    \includegraphics[width=0.50\textwidth]{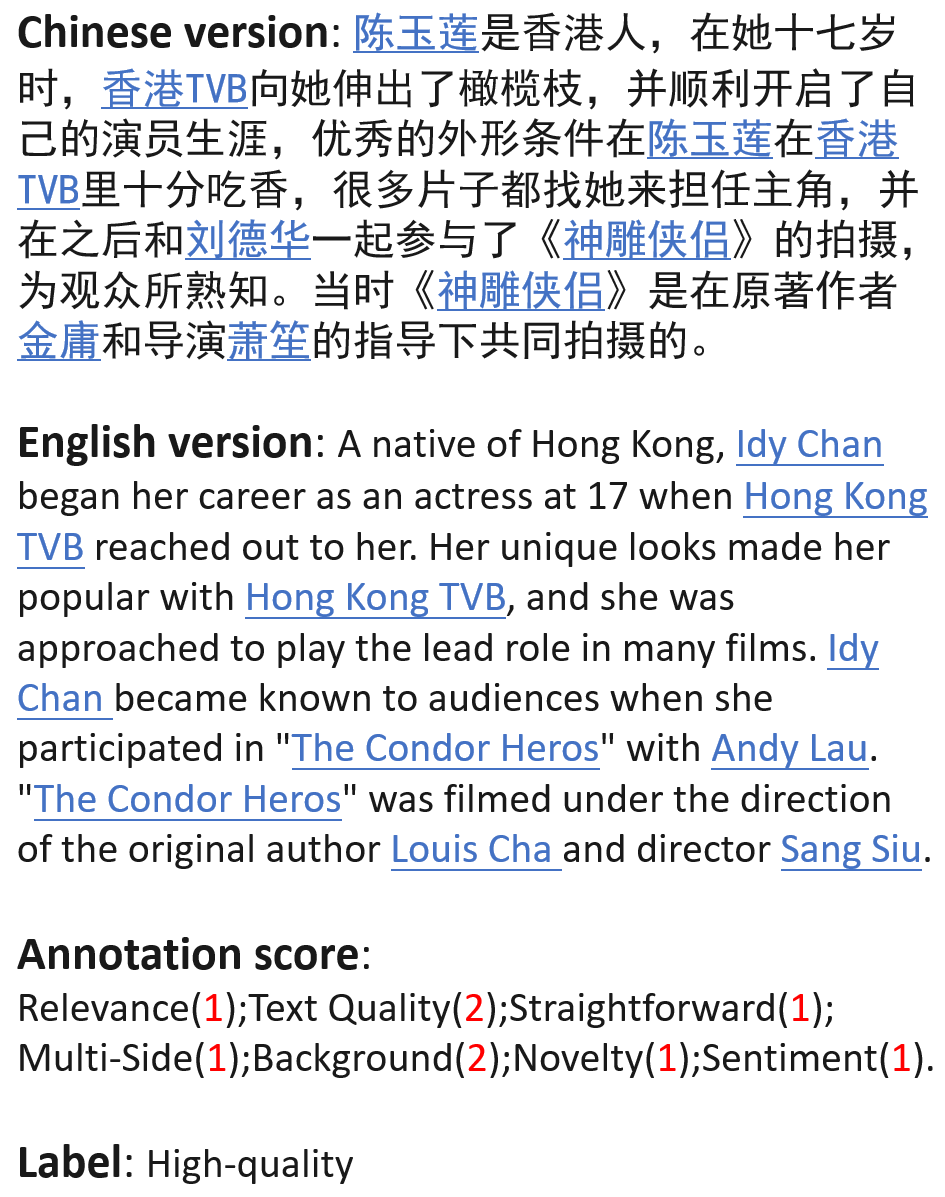} 
    \caption{
An example of our annotation articles where \textcolor{blue}{blue} denotes the entities in the  Baidu Encyclopedia.
\textcolor{red}{red} denotes the scores we annotate in 7 dimensions.
} 
    \label{ner} 
\end{figure}

High-quality articles need to satisfy many characteristics.
Through analysis and surveys, we design 7 relatively objective indicators to annotate the quality of articles.
We randomly collect over 10k articles from Baijiahao \footnote{\url{https://baijiahao.baidu.com/}}  and invite professional annotators to annotate.
Taking Figure~\ref{ner} as an example, we annotate the article in 7 dimensions and then identify quality of article based on the scores of these indicators.
We find it difficult to judge quality of the article from the context alone without any external knowledge, such as ``\begin{CJK}{UTF8}{gbsn}神雕侠侣\end{CJK} (The Condor Heroes)'' is a TV series and ``\begin{CJK}{UTF8}{gbsn}刘德华\end{CJK} (Andy Lau)'' is a famous actor.
Baidu Encyclopedia \footnote{\url{https://baike.baidu.com}} is an online encyclopedia from Baidu that covers almost every known field of knowledge.
In response, we adopt it as the external knowledge base to enhance article identification.
Firstly, we apply Baidu Lexer \footnote{\url{https://ai.baidu.com/ai-doc/NLP/tk6z52b9z}} to extract entities in each article. 
As shown in Figure~\ref{ner}, we extract entities such as ``\begin{CJK}{UTF8}{gbsn}神雕侠侣\end{CJK} (The Condor Heroes)'' and ``\begin{CJK}{UTF8}{gbsn}陈玉莲\end{CJK} (Idy Chan)''.
Secondly, we search each entity in Baidu Encyclopedia, then link these entities to corresponding pages and collect some description information such as English name and some information about the co-occurring entities in the pages.
The Baidu Encyclopedia details of ``\begin{CJK}{UTF8}{gbsn}神雕侠侣\end{CJK} (The Condor Heroes)'' and ``\begin{CJK}{UTF8}{gbsn}陈玉莲\end{CJK} (Idy Chan)'' please refer to Figure~\ref{graph1}.
Consequently, we combine labeled articles with corresponding Baidu Encyclopedia information and propose the \textbf{K}nowledge \textbf{E}nhanced article \textbf{Q}uality \textbf{I}dentification (KE-QI) dataset.

Traditional text classification models such as BERT \cite{bert}, RoBERTa \cite{roberta}, and XLNet \cite{DBLP:journals/corr/abs-1906-08237} only consider the semantic information of the articles themselves, which have certain limitations in identifying article qualities.
Besides, recent works demonstrate that incorporating external knowledge into language models can enhance the representation of language models \cite{sun2019ernie,peters-etal-2019-knowledge,liu2019kbert,10.1162/tacl_a_00360}.
Inspired by these, we first construct the entity graphs from the co-occurrence of entities in articles and Baidu Encyclopedia pages.
We select entity graph instead of knowledge graph because entity graph can apply to any articles and can effectively capture novelty information that knowledge graph cannot express.
We then propose a compound model which leverages Graph Convolutional Network \cite{kipf2016variational} to combine external knowledge from entity graphs for article quality identification.
The model fuses the inner semantic information encoded by a text encoder and external knowledge information encoded by a graph encoder via the gate unit to identify the quality of the article.
In addition, we find that suitable node initialization methods can drastically improve the performance of the graphic encoder.
Therefore, we collect 10 million additional articles from Baijiahao and construct entity graphs based on Baidu Encyclopedia.
We then pre-train the entity representation model of Baidu Encyclopedia (BE-Node2Vec) based on Node2Vec \cite{node2vec} to initialize nodes in the graph encoder.

We conduct extensive experiments on KE-QI.
The results show that our compound model gets the best performance compared with the other baselines, reaching 78\% $F_1$ score on testset of KE-QI.
Moreover, using BE-Node2Vec to initialize graph nodes can significantly help the model to learn more factual knowledge information.

The contributions of our work can be summarized as follows:
\begin{itemize}
    \item To our knowledge, we annotate the first large-scale article quality dataset\footnote{\url{https://aistudio.baidu.com/aistudio/competition/detail/255/0/introduction.}}.
    \item We propose a compound model, which combines semantic and external knowledge information, to enhance  article quality identification.
    \item We publish the well-pretrained entity embeddings BE-Node2Vec, which should be able to help other knowledge related tasks.
\end{itemize}

\section{Related Work}

\paragraph{Text Classification Models} 
Text classification is a classical problem in natural language understanding (NLU).
Typical tasks include sentiment analysis, news categorization, topic classification and natural language inference (NLI) \cite{DBLP:journals/corr/abs-2004-03705}.
With the continuous development of deep learning, a series of neural network models are applied to text classification tasks, including RNN \cite{DBLP:journals/corr/WanLGXPC15,ijcai2017-579}, CNN \cite{DBLP:journals/corr/LeCD17,GUO2019366}, Graph Neural Networks (GNNs) \cite{DBLP:journals/corr/abs-1809-05679,lin-etal-2021-bertgcn} and pre-trained models such as BERT \cite{bert} and RoBERTa \cite{roberta}.
The above models only consider semantic information and ignore factual knowledge information.
In contrast, our approach can combine two types of information to enhance the accuracy of text classification.

\paragraph{Graph Neural Networks}

Graph Neural Networks (GNNs) have shown great potential in tackling graph analytic problems, such as node classification \cite{DBLP:journals/corr/abs-1905-00067,DBLP:journals/corr/abs-1906-02319}, graph classification \cite{DBLP:journals/corr/abs-1905-05178,Zhang_Cui_Neumann_Chen_2018}, link prediction \cite{kipf2016variational,DBLP:journals/corr/abs-1906-04817}, and recommendation \cite{DBLP:journals/corr/abs-1806-01973,DBLP:journals/corr/abs-1902-07243}.
Graph Convolutional Network (GCN) \cite{kipf2016variational} is a further extension while researching on GNNs.
Recently, GCN has been explored in some NLP tasks, such as Text Classification \cite{DBLP:journals/corr/DefferrardBV16,10.1145/3178876.3186005}, Reading Comprehension \cite{DBLP:journals/corr/abs-1905-05460,DBLP:journals/corr/abs-1906-02916}, Machine Translation \cite{bastings-etal-2017-graph,DBLP:journals/corr/abs-1806-09835} and achieve state-of-the-art results in a number of datasets.

\paragraph{Knowledge-Enhanced Language Models}

Recent works show that external knowledge has potential in enhancing a variety of knowledge-intensive NLP tasks.
\citet{sun2019ernie} proposes ERNIE to identify entities mentioned in text and links pre-processed knowledge embeddings of entities to the corresponding positions, which significantly improves various downstream tasks.
Similarly, KnowBert \cite{peters-etal-2019-knowledge} incorporates integrated entity linkers and jointly train them with self-supervised language modeling objective in an end-to-end multi-task model.
In addition,  there are some works that try to inject fixed knowledge into language model, such as K-BERT \cite{liu2019kbert}, KEPLER \cite{10.1162/tacl_a_00360}, KMLMs \cite{DBLP:journals/corr/abs-2111-10962}.

\section{Construction of KE-QI}

To the best of our knowledge, article quality identification is a significant challenge because of the lack of datasets in this area and the absence of suitable criteria to determine the quality of articles.
In response, we propose \textbf{K}nowledge \textbf{E}nhanced article \textbf{Q}uality \textbf{I}dentification dataset(KE-QI).
In this section, we introduce the process of KE-QI construction, including data collection, data annotation, and the corresponding Baidu Encyclopedia pages collection. 

\subsection{Data Annotation} \label{3.2}

We randomly collect over 10k candidate articles from Baijiahao for data annotation.
Through analysis and surveys, we design 7 objective indicators in the range of $\{0,1,2\}$ to annotate the quality of articles.
More details of our annotation rules are shown in Appendix~\ref{sec:appendix}.
The 7 indicators can be briefly summarized as follows:

\paragraph{Relevance} 
Relevance is used to describe the degree of relevance between the content of an article and its title.

\paragraph{Text Quality}
Text quality measures the grammar and spelling levels of the article's sentences.

\paragraph{Straightforward}
Straightforward refers to whether the topic of the article is intuitive, for example, by describing it with a lot of data or summarising it in simple terms.

\paragraph{Multi-Sided}
Multi-Sided means that the articles  have multiple perspectives in description, thus making the topics more authoritative.

\paragraph{Background}
Background describes the wealth of real-world contextual information represented in an article.

\paragraph{Novelty} 
Novelty describes whether an article is relevant to current events or whether the author's ideas are innovative.

\paragraph{Sentiment}
Sentiment denotes the emotional orientation of the content of the article.

\begin{figure}[tb] 
    \centering 
    \includegraphics[width=0.5\textwidth]{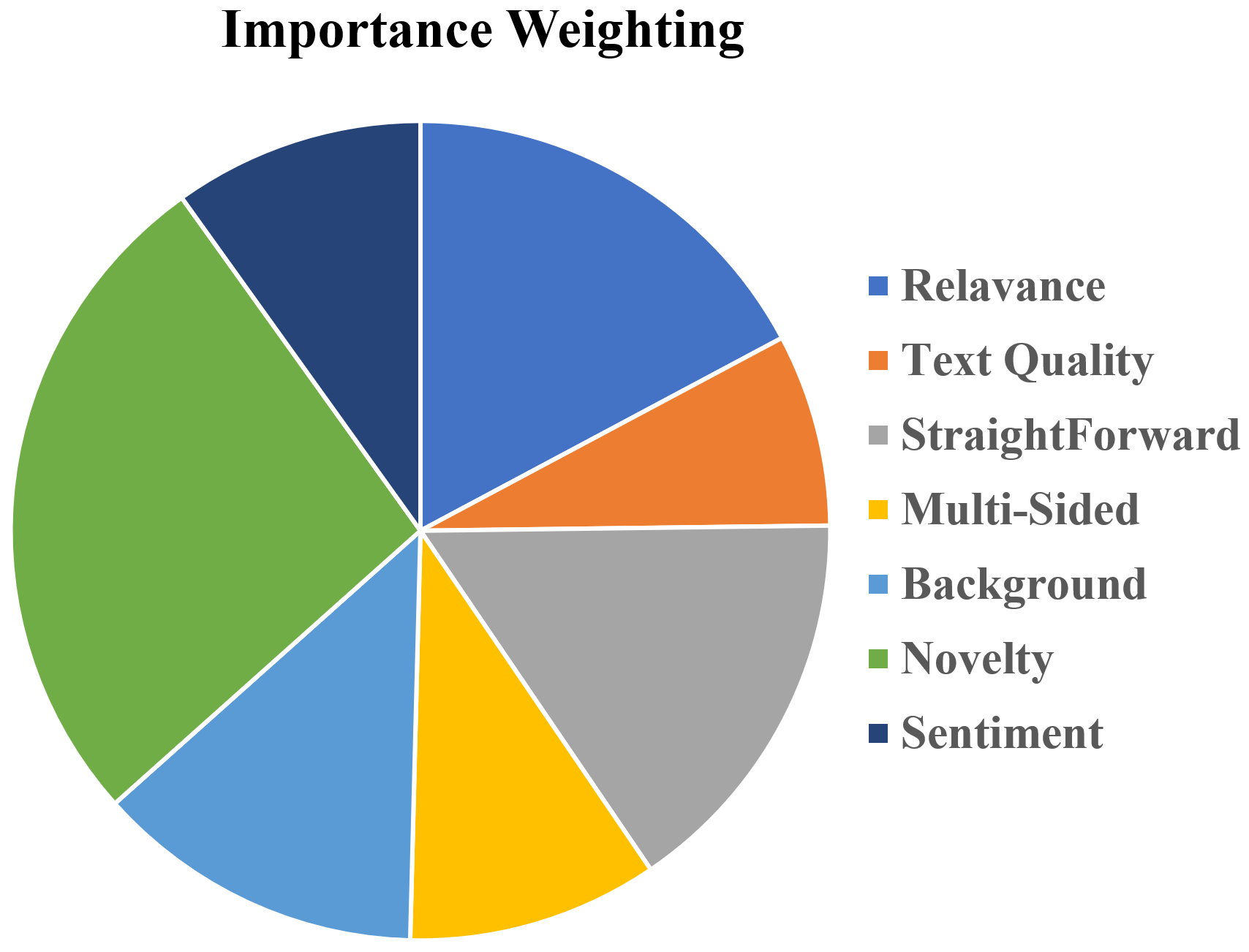} 
    \caption{Weighting of the 7 dimensional indicators in the decision tree model.} 
    \label{weighting} 
\end{figure}

\begin{figure}[tb] 
    \centering 
    \includegraphics[width=0.5\textwidth]{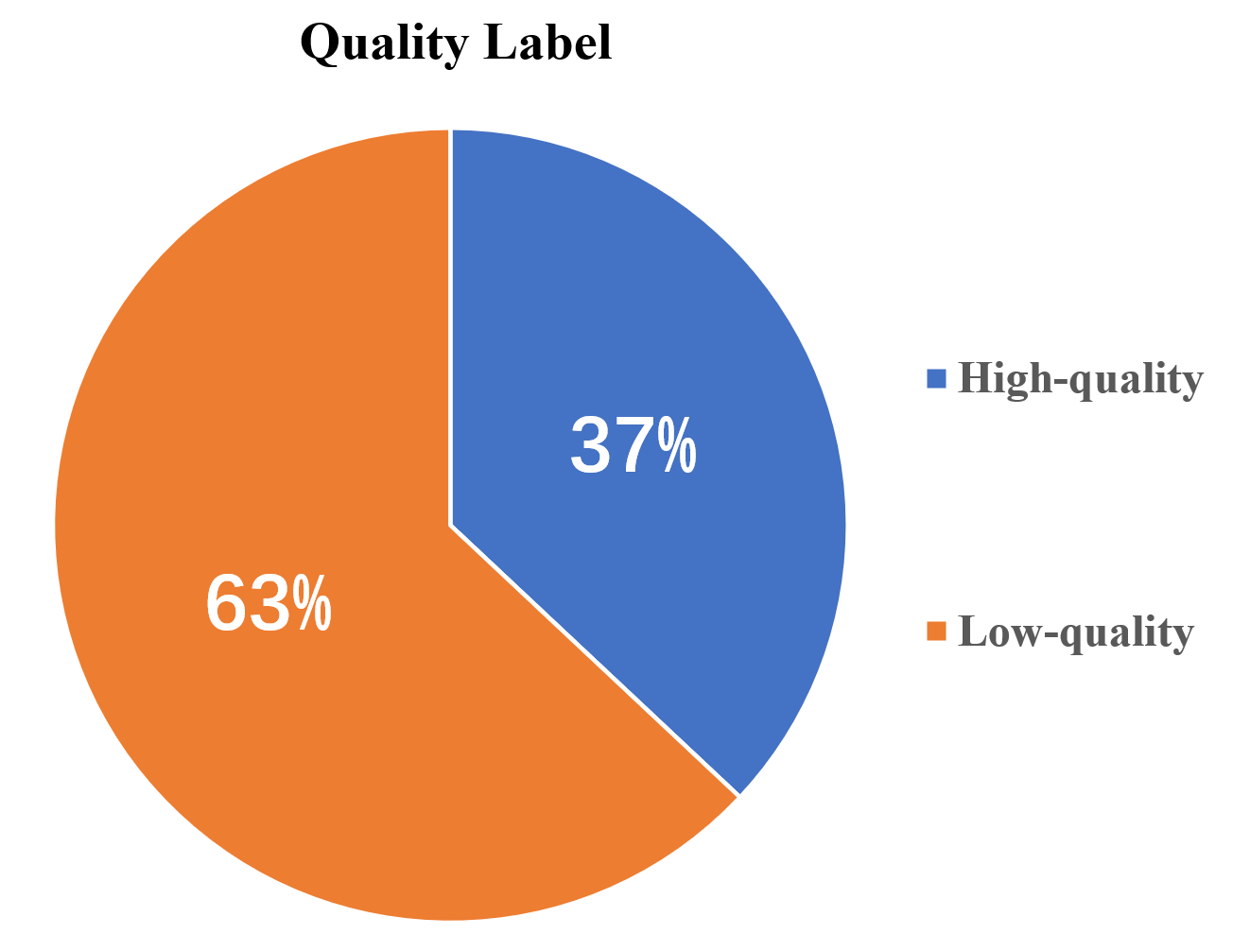} 
    \caption{
    Statistics on the quality of our annotated articles.
} 
    \label{label_dis} 
\end{figure}

A total of four professional data annotators are arranged to annotate these articles.
Testing on 100 articles, the Kappa score~\cite{Fleiss_kappa_2015} of annotators reaches about $0.82$, indicating high consistency.
In order to obtain the final quality labels, we further sample 1000 articles and manually classify them into two categories (high-quality or low-quality).
Then we split 1000 articles and use 900 of them to train a decision tree model to predict classification label by fitting 7 dimensions scores of these articles.
The decision tree model achieves the accuracy of 96$\%$ in the remaining 100 articles.
We finally apply the decision tree model to automatically annotate unlabeled articles in the dataset.
Figure~\ref{weighting} shows contributions of  the 7 dimensions in our trained decision tree model, which indicates that novelty, straightforward and background are the most important indicators compared to the others.
Besides, the distribution of the quality of our annotated articles is displayed in Figure~\ref{label_dis}.
The number of high-quality texts in our annotated datasets is only about 37\%.

\subsection{Baidu Encyclopedia Information Collection}

\begin{table}[tb]
\centering
\begin{tabular}{c|c|c|c}
\hline
\centering
Datasets &Count & Avg. Length & Entity Count \\ \hline
Train   & 7,835   & 3,478      & 1,190,417 \\ 
Valid   & 1,078   & 3,463     &  168,042 \\ 
Test    & 1,106   & 3,474     &  166,121 \\ \hline
\end{tabular}
\caption{Statistics of our KE-QI dataset.}
\label{data}
\end{table}

\begin{figure}[tb] 
    \centering 
    \includegraphics[width=0.5\textwidth]{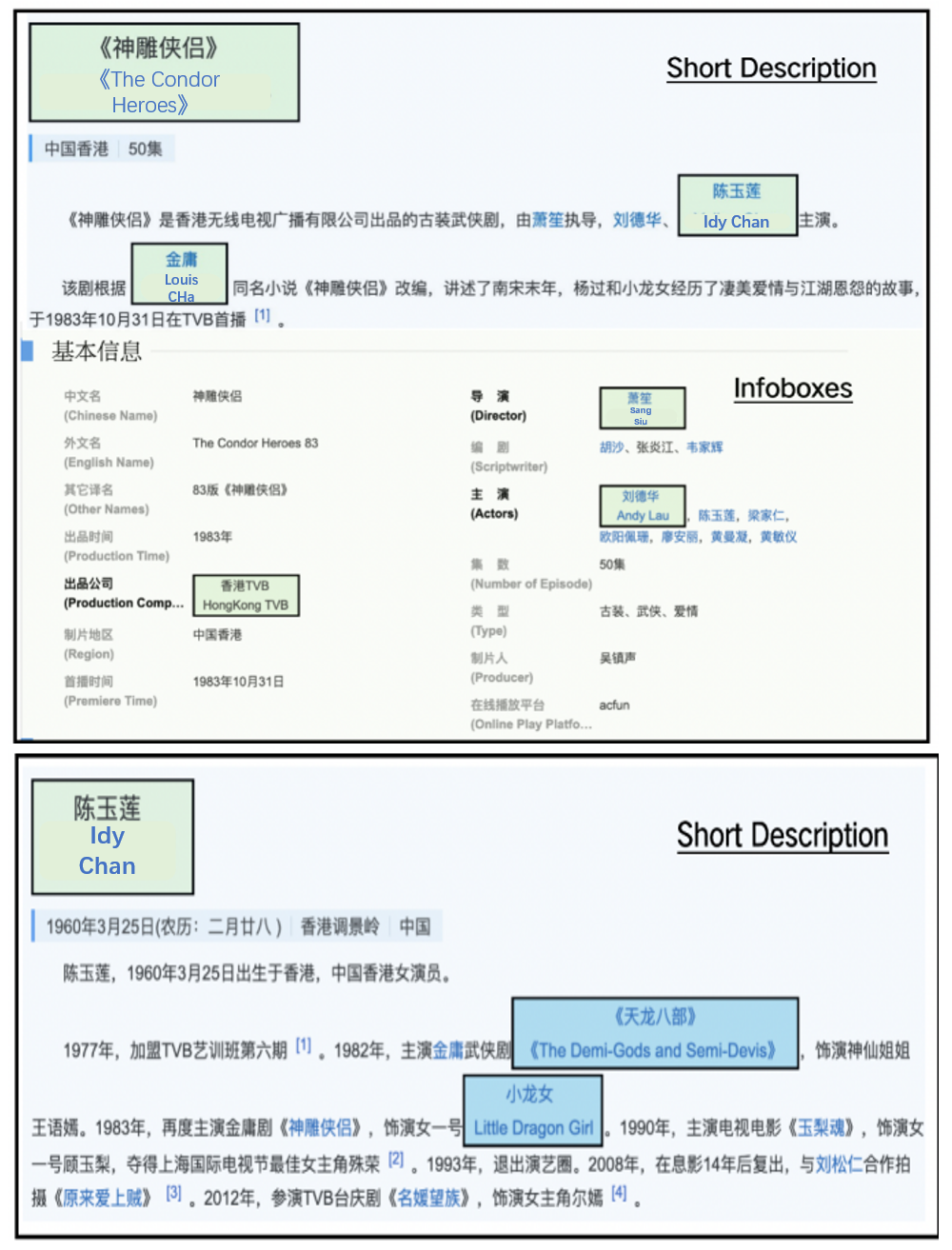} 
    \caption{
The figure shows the Short Description and Infoboxes structures of ``\begin{CJK}{UTF8}{gbsn}神雕侠侣\end{CJK} (The Condor Heroes)'' and ``\begin{CJK}{UTF8}{gbsn}陈玉莲\end{CJK} (Idy Chan)'' in Baidu Encyclopedia pages, where \textcolor{blue}{blue} denotes the entities with hyperlinks.  
}
    \label{graph1} 
\end{figure}

Through analysis, we find that many characteristics of high-quality articles (e.g., multi-sided, background) rely more on extensive external knowledge than inner semantic information of articles.
To provide the model with general knowledge, we extract the entities of articles and then link them to corresponding Baidu Encyclopedia pages and collect some information.

\paragraph{Entity Extraction}

As we all know, terms can be extracted by some Named Entity Recognition (NER) tools.
Among different NER tools, Baidu Lexer can locate the characteristic or property of a particular word and can find at most 38 types of named entities, such as person names with 154 types of sub-categories (e.g., actor, director, etc.).
Taking Figure~\ref{ner} as an example, we adopt Baidu Lexer to extract entities such as  ``\begin{CJK}{UTF8}{gbsn}陈玉莲\end{CJK} (Idy Chan)'' and  ``\begin{CJK}{UTF8}{gbsn}神雕侠侣\end{CJK} (The Condor Heroes)'' in article.

\paragraph{Entity Information Collection}
We use the Baidu Encyclopedia to collect some entity description information such as Chinese name and entities (both entities co-occur in Baidu Encyclopedia page). 
For example, in the Baidu encyclopedia page for ``\begin{CJK}{UTF8}{gbsn}神雕侠侣\end{CJK} (The Condor Heroes)'' shown in Figure~\ref{graph1},
we collect some description information such as Chinese name and English name of ``\begin{CJK}{UTF8}{gbsn}神雕侠侣\end{CJK} (The Condor Heroes)''.
Besides, we annotate the entities such as ``\begin{CJK}{UTF8}{gbsn}陈玉莲\end{CJK} (Idy Chan)'' and ``\begin{CJK}{UTF8}{gbsn}刘德华\end{CJK} (Andy Lau)'' that appear in both the article and the Baidu encyclopedia interface as co-occurrence entities.

Table~\ref{data} shows the statistics of KE-QI.
We then connect articles with corresponding entity links to form the \textbf{K}nowledge \textbf{ E}nhanced article \textbf{Q}uality \textbf{I}dentification dataset (KE-QI).

\section{Proposed Article Quality Classifier} \label{pc}

\begin{figure*}[htb]
    \centering
    \includegraphics[width=\linewidth]{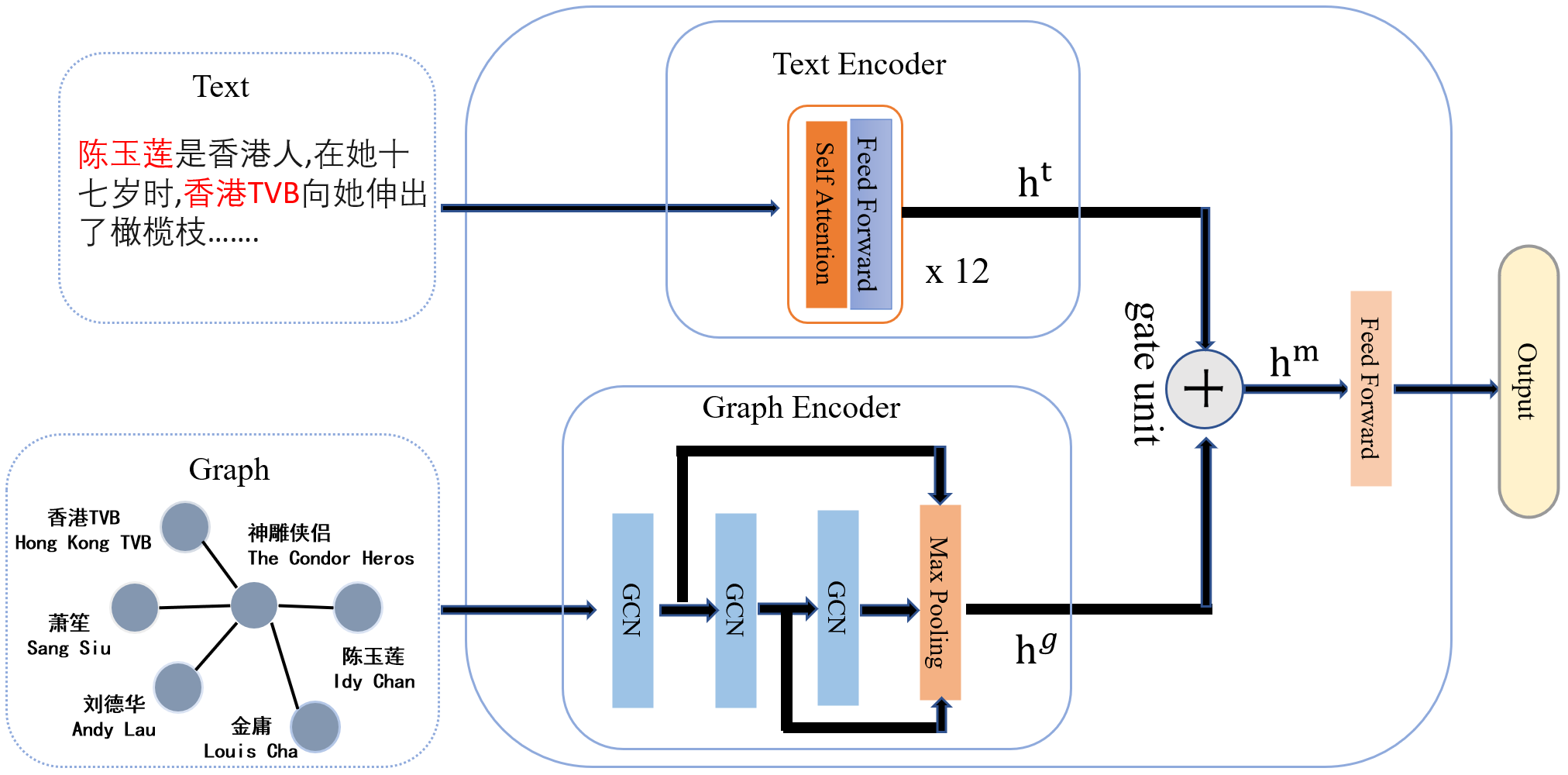}
    \caption{Illustration of our compound model.
    It consists of a semantic encoder module, a graph encoder module, and a gate unit.}
    \label{model}
\end{figure*}

\begin{figure}[tb] 
    \centering 
    \includegraphics[width=0.5\textwidth]{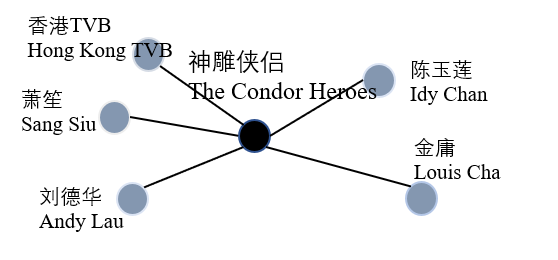} 
    \caption{
    An example of entity co-occurrence graphs constructed by First-Order.
} 
    \label{fo1} 
\end{figure}

Considering both context and novelty of articles, we construct entity co-occurrence graphs for each article in KE-QI to incorporate external knowledge.
Taking Figure~\ref{fo1} as an example, any two connected entities in the graph indicate that they appear together in the article and Baidu Encyclopedia.
Based on entity co-occurrence graphs, 
we then propose a compound model to identify whether an article is high-quality or not by combining inner semantic and external knowledge information.
Figure~\ref{model} shows the structure of our proposed model.
The model contains a graph encoder, a text encoder, and a gate unit to combine the two encoders.
In the following description, we use $x^t$ for the article, $x^{g_v}$ and $x^{g_e}$ for the nodes and edges of the entity co-occurrence graph constructed from the corresponding article $x^t$ based on Baidu Lexer and Baidu Encyclopedia, $y\in\{0,1\}$ for the target, where `1' means high-quality and `0' means low-quality.

\subsection{Entity Co-Occurrence Graph Construction} \label{FO_c}

Inspired by \citet{huang-etal-2019-text}, we design a kind of connection named First-Order to construct entity co-occurrence graph for each article in KE-QI.
First-Order directly describes the entities that co-occur in both the article and the Baidu encyclopedia.
As shown in the Figure~\ref{graph1}, ``\begin{CJK}{UTF8}{gbsn}陈玉莲\end{CJK} (Idy Chan)'' and ``\begin{CJK}{UTF8}{gbsn}神雕侠侣\end{CJK} (The Condor Heroes)'' co-occur in the Baidu encyclopedia pages shown in Figure~\ref{graph1} and article shown in Figure~\ref{ner}, so we construct an edge between two entities.
In contrast, ``\begin{CJK}{UTF8}{gbsn}陈玉莲\end{CJK} (Idy Chan)'' and ``\begin{CJK}{UTF8}{gbsn}胡沙\end{CJK} (Sha Hu)'' have no co-occurrence relationship in article, so there are no edge to connect the two entities.
Entity co-occurrence graphs contain a abundance of external knowledge information.

\subsection{Text Encoder}

Following the common practice, we adopt BERT \cite{bert} as our  backbone text encoder to get semantic representation.
\begin{equation}
    h^t = BERT (x^{t}).
\end{equation}
Like \citet{DBLP:journals/corr/abs-2101-10642}, we just use the hidden state at the sentence tag ($[cls]$) to represent the entire input.

\subsection{Graph Encoder} \label{ge}

Inspired by \citet{DBLP:journals/corr/abs-1911-03082}, we apply GCN \cite{gcn} to represent graph $\{x^{g_v},x^{g_e}\}$ which is constructed  based on its linked text by Baidu Encyclopedia.
The GCN requires an initial node representation, and we try various initialization methods in our experiments.

\begin{equation}
    h^{(0)} = Embedding(x^{g_v}).
\end{equation}
where $x^{g_v}$ denotes nodes of entity graphs in KE-QI, $h^{0}$ denotes representation of $x^{g_v}$.
We then use GCN with $\lambda$ layers to learn the adjacency relations on the graph, and  we empirically determine $\lambda$ as 3 in our experiment.
\begin{align}
    h^{(i)} &= ReLU (GCN ( h^{(i-1)}, x^{g_e})).
\end{align}
where $h^{(i)},i\in \{1,2,3\}$ denote the output of i-th GCN, and we adopt ReLU \cite{pmlr-v15-glorot11a} as activation function for GCN.
Subsequently, we perform a global max-pooling for all outputs of the GCN.
\begin{equation}
    h^{g} = MaxPooling (h^{(1)},h^{(2)},h^{(3)}).
\end{equation}

\subsection{Gate Unit Module}

Inspired by \citet{xu-etal-2021-read}, we build a gate unit to fuse semantic and knowledge information to identify the quality of article.
Given the hidden states $h^t$ and $h^g$ computed in the  above text encoder and graph encoder modules, we propose a gate unit to merge them into a gate unit.
\begin{align}
    g^t &= \sigma (W^t \cdot h^t+b^t), \\
    g^v &= \sigma (W^g \cdot h^g+b^g), \\
    h^m &= g^t \cdot h^t + g^v \cdot h^v.
\end{align}
where $W^t$, $b^t$, $W^g$, $b^g$ are learnable parameters, $\sigma$ is sigmoid function.
Finally, the classification is done by a feed-forward network (FFN).
\begin{align}
    &p (\hat{y}|x^{t}, x^{g_v}, x^{g_e}) \\= 
    &softmax(FFN(h^m)).
\end{align}
Notably, we use softmax rather than sigmoid for this binary classification task as we compare the two methods, and the results show that softmax is $1.8\%$ higher than sigmoid.
The learning process is driven by optimizing the objective. 
\begin{equation}
    \mathcal{L} = - \sum_{i} log p (\hat{y_i} = y_i | x_i^{t}, x_i^{g_v}, x_i^{g_e}).
\end{equation}

\section{BE-Node2Vec Pre-Training} \label{graph_representation}

\begin{figure}[tb] 
    \centering 
    \includegraphics[width=0.5\textwidth]{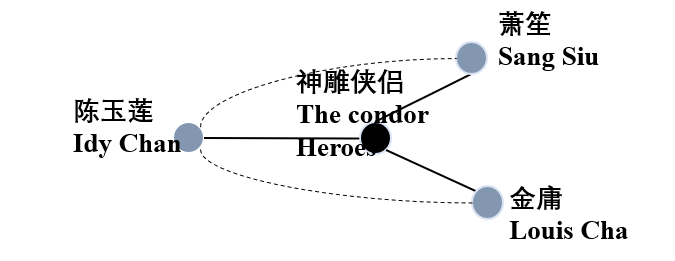} 
    \caption{
    An example of Second-Order graphs, where dotted lines denote the edges of Second-Order connection.
} 
    \label{fo2} 
\end{figure}

As we all know, better initialization method leads to faster convergence and better performance of the neural network.
Inspired by Node2Vec~\cite{node2vec}, we pre-train entity representation model of Baidu Encyclopedia (BE-Node2Vec) to initialize the representation of entities in graph encoder at Section~\ref{ge}. 
Node2Vec  is capable of learning long-distance reliance between terms through random walks.
To preserve long-distance dependent information in entity co-occurrence graphs, we design Second-Order connection based on First-Order connection in Section~\ref{FO_c}.
On the basis of First-Order, Second-Order connects any two nodes in the entity co-occurrence graph with a distance of 2.
Figure~\ref{fo2} is an example for Second-Order.
The distance between ``\begin{CJK}{UTF8}{gbsn}陈玉莲\end{CJK} (Idy Chan)'' and ``\begin{CJK}{UTF8}{gbsn}金庸\end{CJK} (Louis Cha)'' is 2,  because they are connected through ``\begin{CJK}{UTF8}{gbsn}神雕侠侣\end{CJK} (The Condor Heroes)''.
Consequently, we construct dotted line between ``\begin{CJK}{UTF8}{gbsn}陈玉莲\end{CJK} (Idy Chan)'' and ``\begin{CJK}{UTF8}{gbsn}金庸\end{CJK} (Louis Cha)''.
We build entity graph consisting of First-Order and Second-Order for every one in 10 million articles collected from Baijiahao and then pre-train entity representation model of Baidu Encyclopedia (BE-Node2Vec) with the graphs.

\subsection{Relation Weight Design of First-Order and Second-Order}

With the construction of First-Order and Second-Order, the entity graphs contain a vibrant source of information.
Besides, we find that First-Order is more sensible than Second-Order because it directly manifests the connection between entities.
Consequently, we weak the level of importance of Second-Order to differentiate two methods.
Let $n_i$ denotes the i-th node in the walk, $n_{i+1}$ is a node connected with $n_i$.
Relation weight $R_{weight}$ includes $R_{First-Order}$ and $R_{Second-Order}$.
$R_{weight}$ is generated by the following equations while perform Node2Vec dynamically.

\begin{align}
\small
    R_{weight} =
        \begin{cases}
           Count_{n_i,n_{i+1}}\times \alpha,   & \text{FO} \\
           Count_{n_i,n_{i+1}}\times ( 1 - \alpha),   & \text{SO}
        \end{cases}
\end{align}
where FO denotes First-Order, SO denotes Second-Order, and $\alpha$ is 
hyper-parameter set to 0.8 to describe the weight of First-Order and Second-Order connections in our experiment.
To be noticed, the value of the First-Order should always be larger than the Second-Order.
$Count_{n_i,n_{i+1}}$ denotes the number of linked edges between $n_i$ and $n_{i+1}$ in 10 million articles.

\subsection{Entity Representation Learning}

Based on relation weight of two different types of connections, we use the method based on Node2Vec to train BE-Node2Vec model on entity graphs of 10 million articles.

\begin{align}
    & N_{embedding} \\=
    & Node2Vec (\hat{V},\hat{E},R_{weight},l).
\end{align}
where $\hat{V}$ denotes entities extracted from a article, $\hat{E}$ denotes the edges consisting of First-Order and Second-Order, and $l$ is a fixed length of random walk which we set to 10.
BE-Node2Vec can learn the rich structural information of the entity connection graph, which serves as a way to initialize embeddings of nodes in graph encoder to enhance high-quality article identification.

\section{Experiment}

\subsection{Baseline}

\begin{table*}[tb]
\centering
\small
\setlength{\abovecaptionskip}{0.15cm}%
\setlength{\belowcaptionskip}{-0.1cm}%
\begin{tabular}{l|cccc|cccc}
\hline
\textbf{Model}                                                           
                 & \multicolumn{4}{c|}{\textbf{dev}}           &  \multicolumn{4}{c}{\textbf{test}}                        \\ 
\multicolumn{1}{c|}{}                  & \multicolumn{1}{c}{accuracy}              & \multicolumn{1}{c}{precision}       & \multicolumn{1}{c}{recall}             & \multicolumn{1}{c|}{$F_1$}              & \multicolumn{1}{c}{accuracy}              & \multicolumn{1}{c}{precision}       & \multicolumn{1}{c}{recall}             & \multicolumn{1}{c}{$F_1$} 

\cr
\hline
BERT                               & 0.774                        & 0.709                        & 0.725                        &  0.717 &  0.779 & 0.674                        & 0.694                        & 0.684                        \cr \hline 
ERNIE                                   & 0.751                        & 0.722                        & 0.603                        & 0.657                        & 0.765                        & 0.683                        & 0.597                        & 0.637                        \cr \hline 
K-BERT                                      & 0.774                        & \textbf{0.731}                         & 0.676                         & 0.702                         & 0.782                        & \textbf{0.696}                       & 0.654                         & 0.675                        \cr \hline 
GCN                                         & 0.788                        & 0.697                        & 0.819                        & 0.753                         & 0.769                        & 0.622                        & 0.845                      & 0.717 \cr 
+ Tencent-Word2Vec                                         & 0.798                        & 0.706                        & 0.838                         & 0.766 & 0.793                      & 0.660                        & 0.827                        & 0.734  \cr
+ BE-Node2Vec                                         & 0.800                        & 0.727                        & 0.862                         & 0.788 & 0.812                      & 0.679                        & 0.864                        & 0.760  \cr \hline
Our Model  & 0.794  & 0.702 & 0.833 & 0.762 & 0.790 & 0.656 & 0.825 & 0.731 \cr
+ Tencent-Word2Vec  & 0.802                        & 0.717                        & 0.826                        & 0.768                        & 0.800                        & 0.671                        & 0.827                        & 0.741 \cr
+ BE-Node2Vec  & \textbf{0.820}                        & 0.721                       & \textbf{0.887}                        & \textbf{0.796}                        & \textbf{0.824}                        & 0.684                        & \textbf{0.908}                        & \textbf{0.781} \cr
w/o Gate  & 0.779	& 0.701	& 0.770	& 0.734	& 0.772	& 0.642	& 0.770	& 0.700 \cr 
\hline
     
\end{tabular}
\caption{Comparison between BERT, ERNIE, K-BERT, GCN, and our compound model.
``+ Tencent-Word2Vec'' and ``+ BE-Node2Vec'' indicate different node initialization methods.
The default is to randomly initialize graph nodes.
``w/o Gate'' denotes the fact that we replace the gate unit with concatenation.}
\label{results}
\end{table*}

We apply a number of baselines listed below:
\begin{itemize}
    \item BERT \cite{bert}:
    Bert is a Masked Language Model that surprisingly performs well in various language understanding benchmarks.
    \item GCN \cite{gcn}: 
    GCN is a graph-based text classification model which constructs graphs on the node vocabulary.
    The structure of GCN we use is same as the graph encoder in Section~\ref{ge}.
    As a baseline, GCN classifies articles simply by learning the external knowledge information represented by the entity co-occurrence graph.
    \item ERNIE \cite{sun2019ernie}: ERNIE improves the content of BERT masks. 
    BERT masks word pieces, but it is easy for the model to predict the content of a mask from information about the word alone without paying attention to some syntactic and semantic information.
    ERNIE improves on the weaknesses of BERT and uses both entity-level and phrase-level mask mechanisms, which forces the model to learn to focus on some syntactic and semantic information.
    \item K-BERT \cite{liu2019kbert}: K-BERT is a model that incorporates knowledge graph in the input process.
    K-BERT first identifies the entity in the sentence and then obtains the triples from the knowledge graph with that entity as the head entity. 
    Besides, K-BERT applies the mask operation to self-attention to avoid introducing knowledge noise.
\end{itemize}
Besides, we make a number of variations on this model, such as replacing the gating unit with common concatenation, multiple initialisation methods.

\subsection{Experiment Setting}

The accuracy, precision, recall, and $F_1$ of our annotated dataset are reported as the evaluation metrics, which are commonly used in text classification tasks.

For our task, we train the multi-modal entailment model with 5 epochs on 1 Amax-5000 GPU with a batch size of 16.
We use the AdamW \cite{loshchilov_adamw_2019} optimizer and initial learning rate 2e-4.
The rate of dropout \cite{DBLP:journals/corr/abs-1207-0580} is set to 0.5.
The pre-trained BERT model utilizes in the experiments is Chinese BERT \footnote{\url{https://huggingface.co/hfl/chinese-bert-wwm-ext}}.
Tencent-Word2Vec \footnote{\url{https://ai.tencent.com/ailab/nlp/en/embedding.html}} a Word2Vec model on Chinese words and phrases.

\subsection{Results Analysis}

As shown in Table \ref{results}, graph-based models generally get better performance in classification than those counterparts without graph method (BERT, ERNIE and K-BERT).
Even under random initialization of graph nodes, GCN model has increased by an average of 5.1\% in $F_1$ score on KE-QI testset.
The huge gap in performance shows the necessity of external knowledge in enhancing quality identification of articles.
Moreover, under the same setting, our compound model exceeds GCN model by 1.4\%.
This increase proves that the combination of inner semantic information and external knowledge can achieve better classification results than simply using only one of them.

As for graph node initialization, we apply two initialization methods and find that BE-Node2Vec surpasses Tencent-Word2Vec by a large margin.
Our compound model with BE-Node2Vec outperforms others and reaches 78.1\% in $F_1$ score, which demonstrates that our proposed BE-Node2Vec can effectively fuse knowledge into model. 

Notably, replacing the gate unit mechanism with concatenation method causes a huge decline in results.
This shows that the gate unit can judge whether to introduce semantic or knowledge information into the mixed representation to get better classification performance.

\subsection{Case Study}

\begin{figure}[tb] 
    \centering 
    \includegraphics[width=0.5\textwidth]{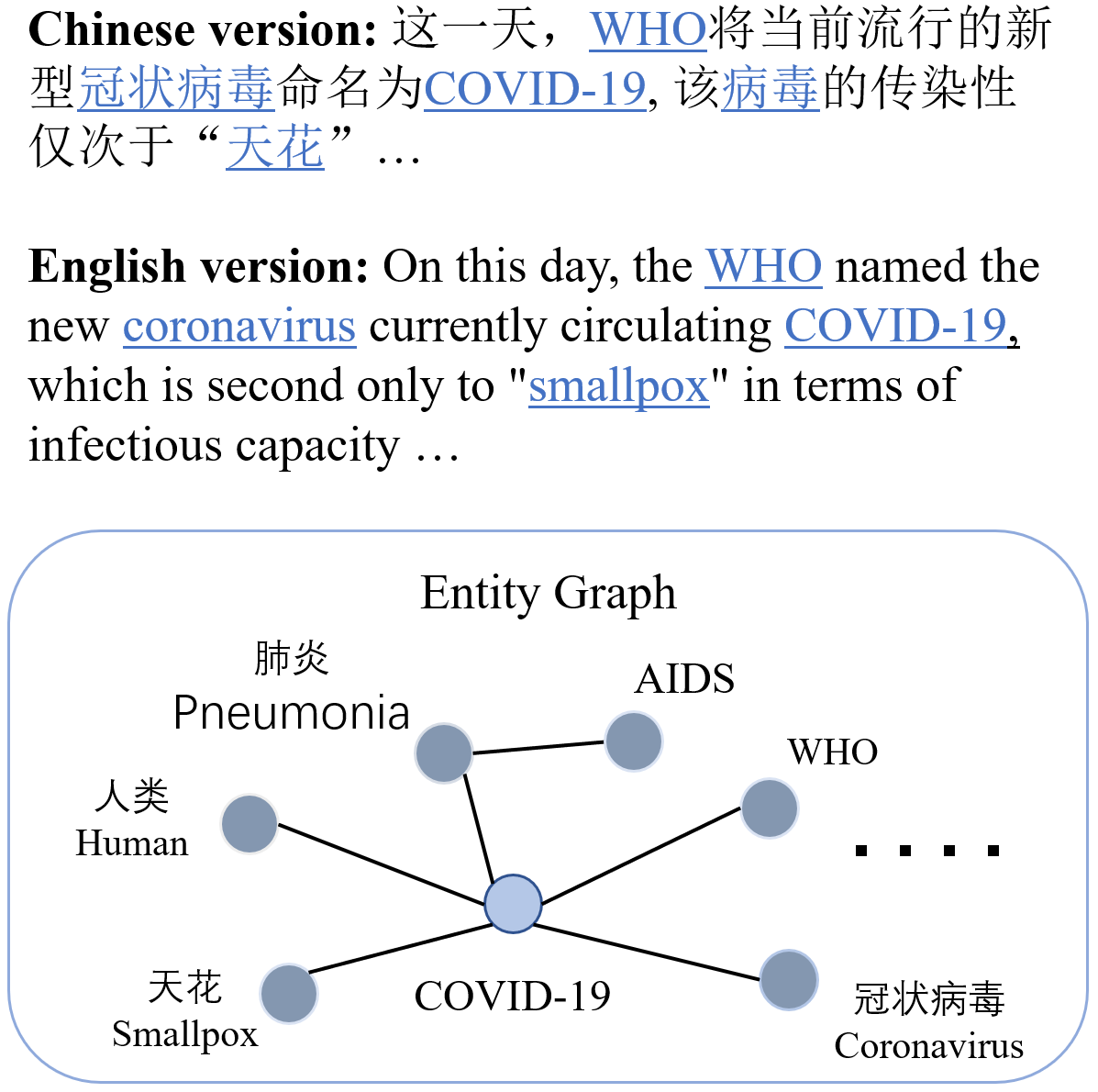} 
    \caption{An example of a dataset cannot be identified by semantic information alone. 
    Our compound model can accurately identify the quality by combining entity graph and content.
} 
    \label{case_study} 
\end{figure}

Figure \ref{case_study} shows a typical example that requires the introduction of external knowledge to identify the quality of articles.
The article narrates the development process of COVID-19 and compares it with other well-known infectious diseases like Smallpox, and it closely follows current events and contains many professional terms.
We annotate this article as high-quality because of its high novelty and  background.
However, BERT fails to identify the article because it only focuses on inner semantic information and does not understand some of the entities in the article, such as WHO and COVID-19.
In contrast, our model does an excellent work of identifying the quality by combining external knowledge and semantic information.
Besides, we measure the distance between COVID-19 and ``\begin{CJK}{UTF8}{gbsn}冠状病毒\end{CJK} (Coronavirus)'' by BE-Node2Vec using cosine similarity, and the result is 0.613, which is consistent with external factual knowledge.

\section{Conclusion}

We propose \textbf{K}nowledge \textbf{ E}nhanced article \textbf{Q}uality \textbf{I}dentification (KE-QI) dataset for article quality identification, which provides a reference for the screening of social media information.
Based on KE-QI, we propose a compound model, which fuse text and external knowledge information via a gate unit to identify the quality of articles.
Besides, we train entities representation model of Baidu Encyclopedia (BE-Node2Vec) to initialize the inputs of our compound model.
Results of the experiments on KE-QI demonstrate the effectiveness of our compound model and BE-Node2Vec.

In future work, we will investigate more effective ways of incorporating external knowledge into language models for article quality identification.
In addition, our dataset currently does not contain information about the images inside the articles, which is a priority issue for our later research.
\bibliography{anthology,custom}
\bibliographystyle{acl_natbib}

\appendix

\section{Example Appendix}
\label{sec:appendix}

\subsection{Data Annotation Rules} \label{data_annotation_details}

\begin{figure*}[tb]
    \centering
    \includegraphics[width=16cm]{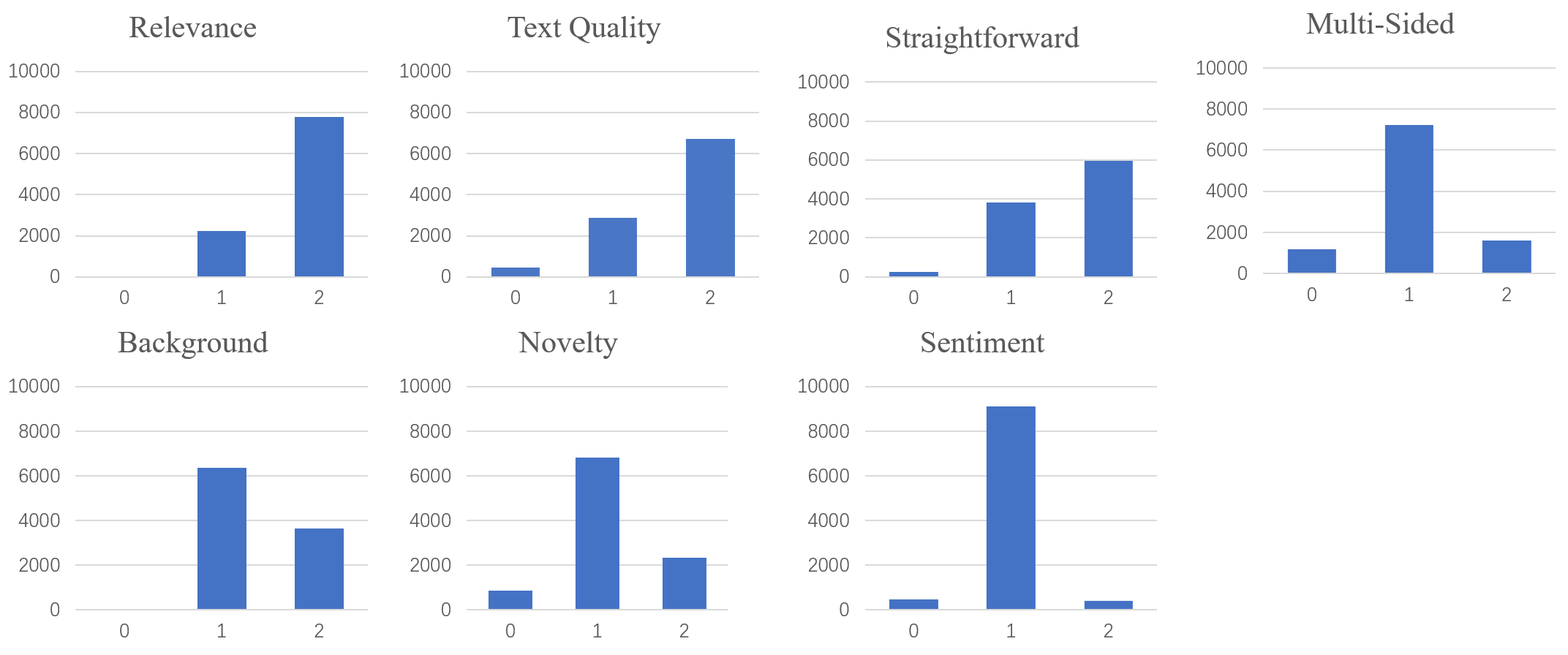}
    \caption{Data distribution for relevance, text quality, straightforward, multi-sided, background, novelty, sentiment.}
    \label{dataset:annotation}
\end{figure*}

Here we introduce the meaning and details of the 7 criteria for identifying high-quality articles as follows.
We manually annotate 7 criteria of an article with scores in the range of $\{0,1,2\}$.
Among them, `0' represents complete in conformity, `1' represents existing but not prominent enough, and `2' represents very prominent and appropriate in corresponding aspect.

\paragraph{Relevance} 
A high-quality article should have an appropriate and non exaggerated title.
We measure the relevance between content and title of an article in order to avoiding deviation.

\paragraph{Text Quality}

We design this indicator to measure grammatical and spelling quality of the sentences in an article.

\paragraph{Straightforward}

Straightforward consists of two aspects:
\begin{itemize}
    \item The article uses a large number of accurate figures or data to portray the matter being written about.
    The article quantifies the whole picture of the matter through data, allowing the user to understand the matter more directly and the content of the article is more accurate.
    \item The article uses clear and concise language to summarise or define the characteristics of the matter being written about so that the user receives the aspects of the matter more directly and the topics are expressed more concisely.
\end{itemize}


\paragraph{Multi-Sided}
This criterion is sufficient if the article meets any of the following characteristics: 
\begin{itemize}
    \item Cite sufficient external quotations (examples, evidence, arguments, etc.) to argue or analyse central ideas or personal opinions, to enhance the persuasiveness of the essay and make it stand up to scrutiny and convince readers.
    \item A rigorous logic of thought that uncovers the multifaceted nature of things (without resorting to external quotes or historical context) and provides a multidimensional presentation of objects and perspectives that enables users to understand the content from different dimensions, making the subject or point explicit and insightful.
\end{itemize}
However, if the essay is not in the style of an argumentative essay, we will award a score as 1.

\paragraph{Background}
We want to have an in-depth article with a clear background to the subject matter, with points and lines of information, such as a timeline, a line of events, etc.
We don't think there is any depth to a straightforward narrative or a simple list of events.

\paragraph{Novelty} 
We define the novelty of an article as the following:
\begin{itemize}
    \item Content topics relate to popular events or extended developments of events and hottest topics of discussion (including but not limited to entertainment, film and television, etc.)
    \item Recent major events at home and abroad or by prominent figures, such as major domestic current affairs news, international conflicts, recent changes in cross-strait relations, etc.
    \item The author has an original and analytical perspective on common events, with novel and rare perspectives, and expresses content that is refreshing and expands the perception.
\end{itemize}

\paragraph{Sentiment}
We believe that insightful articles spread enthusiasm and confidence.
We define positive articles into the following three categories:
\begin{itemize}
    \item The author's perspective is inspiring and motivating.
    \item The atmosphere of the article is inviting, touching humanity, inspiring the people and arousing emotional resonance
    \item Using humorous language to promote the user's sentiment, the content is expressed in a relaxing way to make the user feel comfortable.
\end{itemize}

The distribution of the 7 indicators is shown in Figure~\ref{dataset:annotation}.

\end{document}